# Measuring Massive Multitask Chinese Understanding

**Hui Zeng**
Besteasy (Beijing) Language Technology Co., Ltd.
felix_zeng_ai@aliyun.com

## Abstract

The development of large-scale Chinese language models is flourishing, yet there is a lack of corresponding capability assessments. Therefore, we propose a test to measure the multitask accuracy of large Chinese language models. This test encompasses four major domains, including medicine, law, psychology, and education, with 15 subtasks in medicine and 8 subtasks in education. We found that the best-performing models in the zero-shot setting outperformed the worst-performing models by nearly 18.6 percentage points on average. Across the four major domains, the highest average zero-shot accuracy of all models is 0.512. In the subdomains, only the GPT-3.5-turbo model achieved a zero-shot accuracy of 0.693 in clinical medicine, which was the highest accuracy among all models across all subtasks. All models performed poorly in the legal domain, with the highest zero-shot accuracy reaching only 0.239. By comprehensively evaluating the breadth and depth of knowledge across multiple disciplines, this test can more accurately identify the shortcomings of the models.

## 1 Introduction

With the impressive debut of large models such as ChatGPT(Brockman et al., 2023), a series of Chinese-capable large models, including ChatGLM(Du et al., 2022), MOSS(Sun et al., 2023), Wenxin Yiyan, Tongyi Qianwen, and Shangliang, have been released successively. While there are well-established evaluation methods for large English language models, such as MMLU(Hendrycks et al., 2021), there is still a lack of evaluations for large Chinese language models. Therefore, it is imperative to introduce a scientific evaluation method for large Chinese models and provide high-quality Chinese evaluation datasets. Chinese pre-trained large models based on the Transformer(Vaswani et al., 2017) architecture are trained using massive textual corpora, including Chinese encyclopedic data, a large number of Chinese e-books, and numerous Chinese websites. However, the capabilities of these models in understanding and solving problems across various domains have not been scientifically and comprehensively evaluated.

Since most large language models released recently have undergone prompt-tuning training, we provide both zero-shot and few-shot testing modes, with the latter offering five examples to the models. The test questions are single-choice and multiple-choice, with each question potentially having one or multiple correct answers, making them more akin to human exams and increasing their difficulty.

This test covers four major categories: medicine, law, psychology, and education. Medical questions are drawn from university medical examinations, legal questions from the National Uniform Legal Professional Qualification Examination, psychology questions from psychological counselor exams and graduate entrance exams in psychology, and education questions from the Chinese National College Entrance Examination. The test spans a wide range of disciplines and covers high-level subject knowledge points, making it suitable for assessing the comprehensive capabilities of large models.

It is worth noting that the MOSS model, with as many as 16 billion parameters, demonstrates a zero-shot accuracy of less than 26% across four major domains, making it second to last among all models in the evaluation. In contrast, the GPT-3.5-turbo model, with a minimum of 175 billion parameters, achieved an average zero-shot accuracy of 41.3% (see Table 1 and Figure 1).

The evaluation results indicate that, despite the astonishing progress of large models recently, the most advanced models have not yet reached expert levels in any given domain. The accuracy of all models in legal domain tasks is close to random accuracy. Test codes and test results can be found at github.com/Felixgithub2017/MMCU.

## 2 A Multitask Test

We constructed a large-scale, multi-task test consisting of single and multiple-choice questions from various branches of knowledge. The test encompasses the fields of medicine, law, psychology, and education, with medicine divided into 15 sub-tasks and education into 8 sub-tasks. The questions in the dataset were manually collected by professionals from freely available online resources, including university medical examinations, national unified legal professional qualification examinations, psychological counselor exams, graduate entrance examinations for psychology majors, and the Chinese National College Entrance Examination.

In total, we collected 11,900 questions, which we divided into a few-shot development set and a test set. The few-shot development set contains 5 questions per topic, amounting to 55 questions in total. The test set comprises 11,845 questions.

### 2.1 Medicine

The medical disciplines include basic medical sciences, pharmacology, nursing, pathology, clinical medicine, infectious diseases, surgery, anatomy, medical imaging, parasitology, immunology, pediatrics, dermatovenereology, histology and embryology, and pharmaceutical analysis. The medical domain comprises a total of 2,819 questions.
Here is an example of a medical question:
首次急性发作的腰椎间盘突出的治疗方法首选
A. 绝对卧床休息，3 周后戴腰围下床活动
B. 卧床休息，可以站立坐起
C. 皮质类固醇硬膜外注射
D. 髓核化学溶解

### 2.2 Law

The legal domain questions cover the theories of socialism with Chinese characteristics and the rule of law, jurisprudence, constitutional law, Chinese legal history, international law, judicial system and legal professional ethics, criminal law, criminal procedure law, administrative law and administrative litigation law, civil law, intellectual property law, commercial law, economic law, environmental and natural resources law, labor and social security law, private international law, international economic law, civil procedural law, and judicial system and legal professional ethics. The legal domain consists of 3,695 questions in total.
Here is an example of a legal question:
根据法律规定，下列哪一种社会关系应由民法调整？
A. 甲请求税务机关退还其多缴的个人所得税
B. 乙手机丢失后发布寻物启事称：“拾得者送还手机，本人当面酬谢”
C. 丙对女友书面承诺：“如我在上海找到工作，则陪你去欧洲旅游”
D. 丁作为青年志愿者，定期去福利院做帮工

### 2.3 Psychology

The subjects covered in psychology topics include General Psychology, Personality and Social Psychology, Developmental Psychology, Mental Health and Psychological Disorders, Introduction to Psychological Counseling, Counseling Theories, Psychological Assessment, Basic Counseling Skills, and Counseling Methods. There are a total of 2,001 questions in the field of psychology.
Here is an example of a psychology question:
把与自己本无关系的事情认为有关，这种临床表现最可能出现于：
A. 被害妄想
B. 钟情妄想
C. 关系妄想
D. 夸大妄想

### 2.4 Education

This section comprises Chinese language, mathematics, physics, chemistry, political science, history, geography, and biology. The questions are derived from the National Unified College Entrance Examination in China (Gaokao). The educational domain encompasses a total of 3,331 questions.
Here is an example of a mathematics question:

若圆锥的侧面积等于其底面积的 3 倍，则该圆锥侧面展开图所对应扇形圆心角的度数为（ ）
A. 60°
B. 90°
C. 120°
D. 180°

## 3 Experiment

### 3.1 Experiment Setup

In order to measure the performance of models in multi-task settings, we calculated the zero-shot and few-shot accuracies for all models across all tasks. We evaluated the Bloom series, including bloomz_560m, bloomz_1b1, bloomz_3b, and bloomz_7b1_mt; as well as ChatGLM 6B developed by the Knowledge Engineering Group (KEG) & Data Mining at Tsinghua University, MOSS 16B developed by Fudan University, and OpenAI's GPT-3.5-turbo.

In the zero-shot mode, we feed questions directly into models to obtain answers and calculate accuracy.
Here is an example of a zero-shot prompt:
**请阅读以下选择题并给出正确选项，不要解释原因。**
在平面直角坐标系中，点 P（m－3，4－2m）不可能在（　）
A. 第一象限
B. 第二象限
C. 第三象限
D. 第四象限
**正确答案的序号是：**
In this text, the bold portions represent the prefixes and suffixes of the question. The prefix instructs the model on how to provide an answer, while the suffix guides the model in outputting the English letter representing the answer. In the few-shot mode, we first provide the model with five examples of questions and answers, followed by the question for which the model is expected to generate an answer.

| | bloomz_560m | bloomz_1b1 | bloomz_3b | bloomz_7b1_mt | ChatGLM 6B | MOSS 16B | GPT-3.5-turbo |
|---|---|---|---|---|---|---|---|
| **Medicine** | 0.298 | 0.213 | 0.374 | 0.364 | 0.338 | 0.234 | 0.512 |
| **Law** | 0.163 | 0.140 | 0.180 | 0.174 | 0.169 | 0.133 | 0.239 |
| **Psychology** | 0.201 | 0.187 | 0.319 | 0.346 | 0.288 | 0.211 | 0.447 |
| **Education** | 0.247 | 0.275 | 0.315 | 0.316 | 0.333 | 0.253 | 0.455 |
| **Average** | **0.227** | **0.204** | **0.297** | **0.300** | **0.282** | **0.208** | **0.413** |

Table 1: Zero-shot Accuracy of All Models Across Four Major Domains

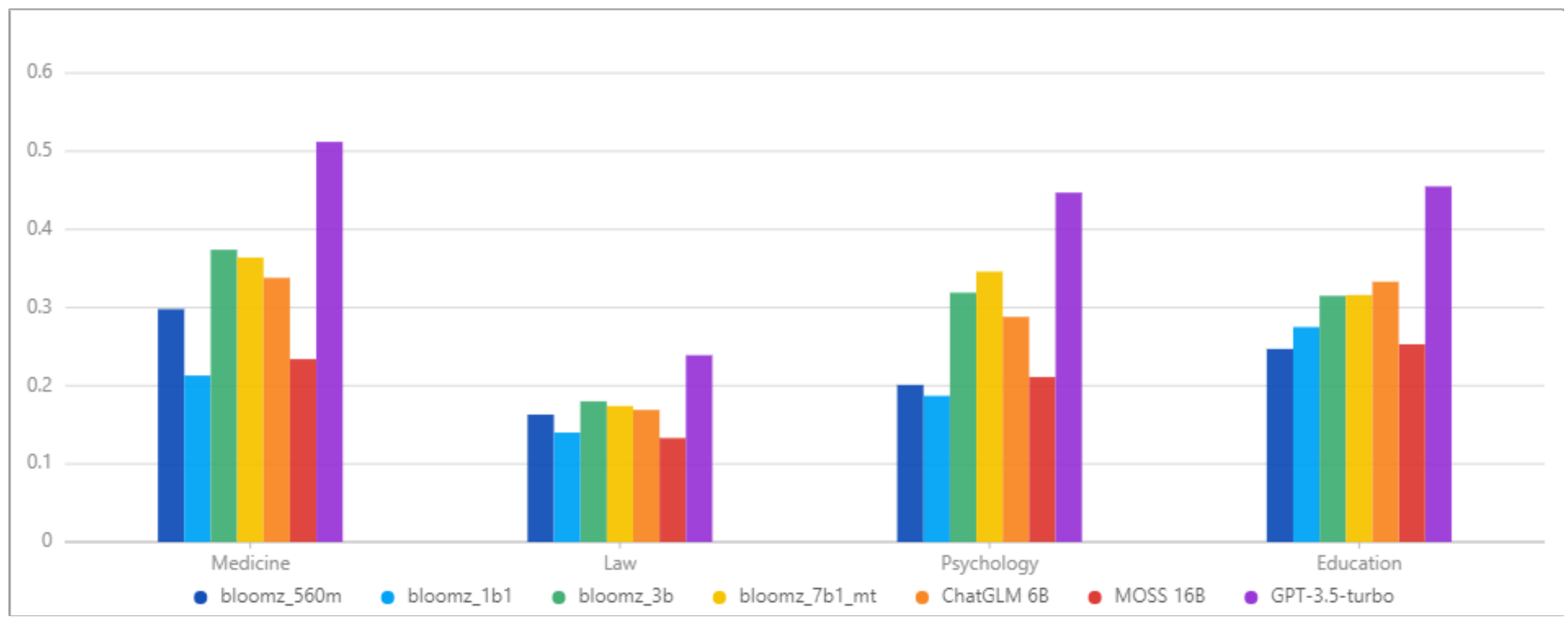

Figure 1: Zero-shot Accuracy of All Models Across Four Major Domains

### 3.2 RESULTS

**Model Scale and Accuracy.** We compare the zero-shot accuracy of various models in Table 1 and Figure 1. We find that GPT-3.5-turbo is far ahead in all four domains. Additionally, we observe that the MOSS 16B model, despite having 16 billion parameters, exhibits an accuracy close to random (approximately 25%). In contrast, the zero-shot accuracy of the lower-parameter Bloom family models, including bloomz_1b1, bloomz_3b, bloomz_7b1_mt, and ChatGLM 6B, is higher. models. These results indicate that while the number of model parameters is a crucial factor for achieving strong performance, the manner and data used for training are also of great importance.

**Subtask Comparison.** Through our tests, we found that GPT-3.5-turbo achieves the highest accuracy in most subtasks, followed closely by ChatGLM 6B. However, the performance of both models is not consistent across tasks. Table 2 and Figure 2 display the accuracy of all models on the educational subtasks. They reveal that both models exhibit accuracy below 60% across all tasks, with GPT-3.5-turbo ranging from 59.9% in biology to 31.0% in Chinese language, and

| | **bloomz_560m** | **bloomz_1b1** | **bloomz_3b** | **bloomz_7b1_mt** | **ChatGLM6B** | **MOSS 16B** | **GPT-3.5-turbo** |
|---|---|---|---|---|---|---|---|
| **Chinese** | 0.233 | 0.283 | 0.248 | 0.205 | 0.256 | 0.233 | 0.310 |
| **Math** | 0.251 | 0.257 | 0.281 | 0.325 | 0.307 | 0.257 | 0.427 |
| **Physics** | 0.173 | 0.208 | 0.185 | 0.202 | 0.256 | 0.208 | 0.327 |
| **Chemistry** | 0.280 | 0.340 | 0.280 | 0.140 | 0.300 | 0.280 | 0.440 |
| **Politics** | 0.239 | 0.255 | 0.329 | 0.401 | 0.329 | 0.268 | 0.545 |
| **History** | 0.279 | 0.296 | 0.421 | 0.432 | 0.448 | 0.245 | 0.513 |
| **Geography** | 0.255 | 0.271 | 0.336 | 0.411 | 0.346 | 0.284 | 0.478 |
| **Biology** | 0.262 | 0.287 | 0.443 | 0.414 | 0.422 | 0.245 | 0.599 |
| **Average** | **0.247** | **0.275** | **0.315** | **0.316** | **0.333** | **0.253** | **0.455** |

Table 2: Zero-shot accuracy of all models on education subtasks

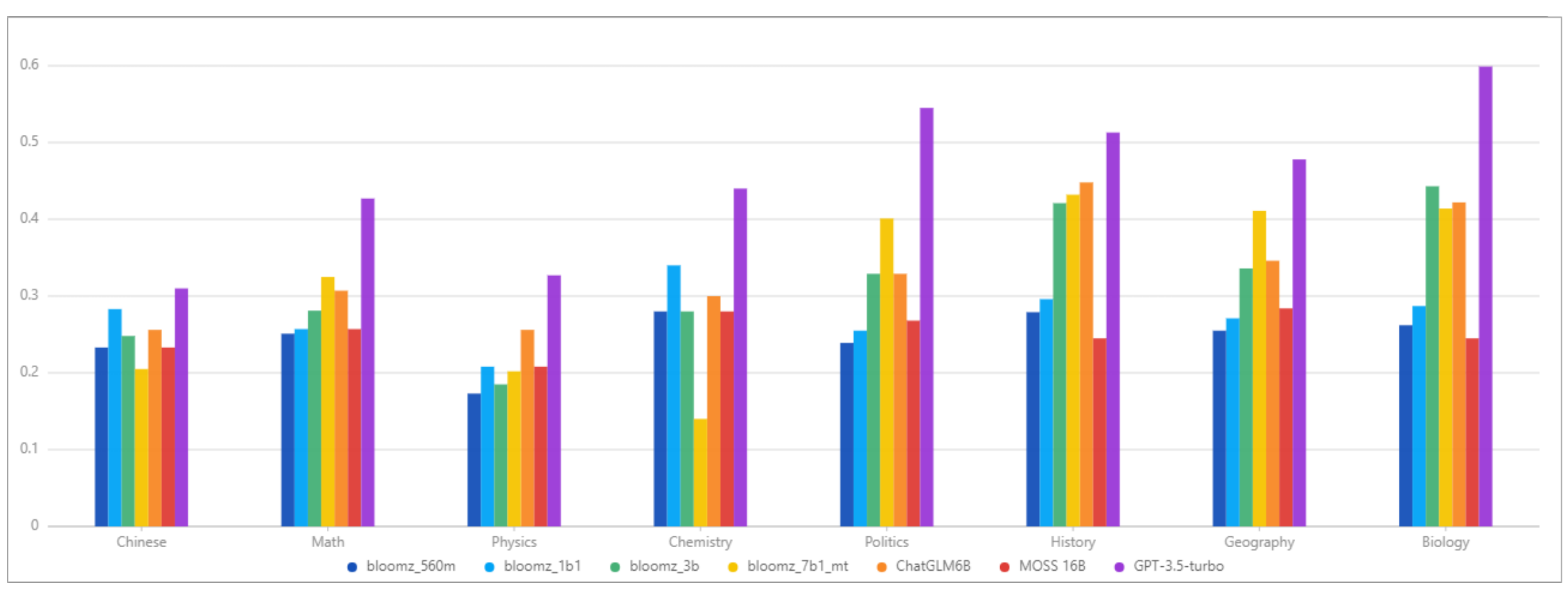

Figure 2. Zero-shot accuracy of all models on education subtasks

Although the bloomz_560m model has the smallest number of parameters, it outperforms the larger-parameter bloomz_1b1 and MOSS 16B ChatGLM 6B ranging from 44.8% in history to 25.6% in physics.

Overall, all models underperform in physics tasks. Table 2 and Figure 2 show that the accuracy of computationally intensive subjects like mathematics and physics tends to be lower. For GPT-3.5-turbo, the tasks with the lowest accuracy are, in order, Chinese language, physics, and mathematics. We speculate that part of the reason is the insufficient Chinese training data for GPT-3.5-turbo, leading to poor performance in Chinese language arts tasks. Additionally, the model tends to solve descriptive questions more easily compared to procedural problems.

Our tests also indicate that all models exhibit varying degrees of performance degradation in the few-shot mode (see Table 4). For example, compared to zero-shot accuracy, GPT-3.5-turbo experiences a decline in few-shot accuracy in the Chinese, chemistry, politics, and geography subtasks. This trend is even more pronounced in the ChatGLM 6B model, where the few-shot accuracy is lower than the zero-shot accuracy for all educational subtasks.

| | bloomz_560m | bloomz_1b1 | bloomz_3b | bloomz_7b1_mt | ChatGLM6B | MOSS 16B | GPT-3.5-turbo |
|---|---|---|---|---|---|---|---|
| **Three basics of medicine** | 0.311 | 0.274 | 0.375 | 0.415 | 0.371 | 0.231 | 0.552 |
| **Pharmacology** | 0.265 | 0.235 | 0.380 | 0.360 | 0.255 | 0.285 | 0.520 |
| **Nursing** | 0.330 | 0.278 | 0.372 | 0.368 | 0.339 | 0.238 | 0.516 |
| **Pathology** | 0.312 | 0.267 | 0.392 | 0.341 | 0.358 | 0.278 | 0.506 |
| **Clinical Medicine** | 0.347 | 0.198 | 0.426 | 0.554 | 0.455 | 0.317 | 0.693 |
| **Infectious Diseases** | 0.295 | 0.242 | 0.398 | 0.460 | 0.401 | 0.254 | 0.587 |
| **Surgery** | 0.365 | 0.251 | 0.397 | 0.374 | 0.361 | 0.279 | 0.525 |
| **Anatomy** | 0.182 | 0.136 | 0.227 | 0.227 | 0.273 | 0.136 | 0.500 |
| **Medical Imaging** | 0.450 | 0.050 | 0.600 | 0.450 | 0.350 | 0.250 | 0.550 |
| **Parasitology** | 0.330 | 0.240 | 0.390 | 0.250 | 0.330 | 0.180 | 0.430 |
| **Immunology** | 0.282 | 0.147 | 0.331 | 0.344 | 0.319 | 0.178 | 0.515 |
| **Pediatrics** | 0.390 | 0.258 | 0.385 | 0.380 | 0.399 | 0.263 | 0.540 |
| **Dermatology** | 0.255 | 0.255 | 0.392 | 0.510 | 0.471 | 0.275 | 0.627 |
| **Histoembryology** | 0.058 | 0.130 | 0.208 | 0.188 | 0.149 | 0.143 | 0.364 |
| **Pharmaceutical Analysis** | 0.292 | 0.236 | 0.333 | 0.236 | 0.236 | 0.208 | 0.250 |
| **Average** | **0.298** | **0.213** | **0.374** | **0.364** | **0.338** | **0.234** | **0.512** |

Table 3: Zero-shot Accuracy of All Models on Medical Sub-tasks

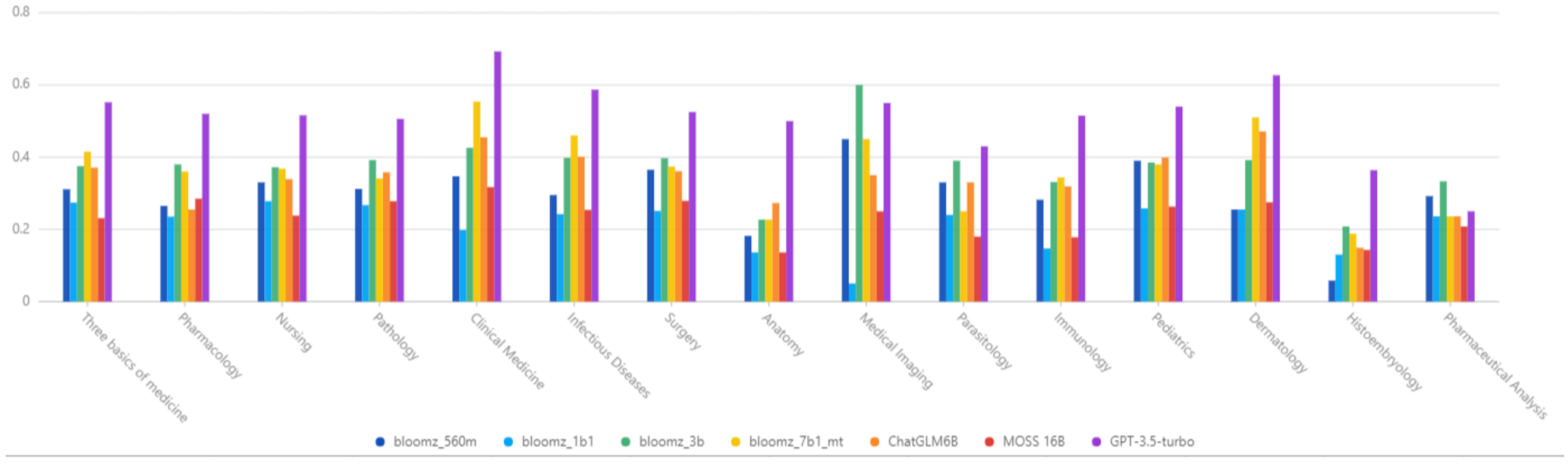

Figure 3: Zero-shot Accuracy of All Models on Medical Sub-tasks

We hypothesize that this may be due to GPT-3.5-turbo and ChatGLM 6B having undergone extensive fine-tuning with instructions and alignment with human preferences. Consequently, the five examples provided in the few-shot mode may inadvertently introduce confusion for the models.

## 4 DISCUSSION

Similar to the English MMLU test, our testing approach does not require large training sets. We assume that the models have already acquired the necessary knowledge by reading vast amounts of diverse text from the internet, a process commonly referred to as pre-training.

Humans primarily learn new knowledge by reading books, listening to lectures, and completing practice exercises. Therefore, we offer a few-shot testing mode and provide development and test sets for each task. The development set is used for few-shot prompts, while the test set is employed to calculate the final accuracy.

We find that there is significant room for improvement in current large-scale Transformer(Vaswani et al., 2017) models. Taking the medical domain illustrated in Table 3 and Figure 3 as an example, the accuracy of most models in multiple medical subtasks is below 60%. The zero-shot accuracy of ChatGLM 6B, which is second only to GPT-3.5-turbo, does not exceed 50% in many medical subtasks.
Therefore, future research should particularly focus on improving model accuracy in vertical domain tasks such as medicine, law, and others. Additionally, none of the models have achieved excellent performance (90%) across all tasks. It remains unclear whether simply increasing the number of parameters could lead to improvements in these tasks. Data might also be a significant bottleneck, as the training of these large models typically relies on massive amounts of publicly available internet data. Efficient filtering of data and annotation of high-quality data in vertical domains are also crucial.
As the APIs for large Chinese models such as Wenxin Yiyan, Tongyi Qianwen, and Shangliang are not yet available, we have been unable to conduct extensive testing. If these models provide access to their APIs, we will evaluate them promptly and update the results in this paper.

## 5 CONCLUSION

We introduce a Chinese language test, encompassing four major domains and numerous subtasks, to evaluate the ability of pre-trained large Chinese language models in solving problems across multiple domains. We find that a larger number of model parameters does not necessarily result in better performance; the training approach and the quality of data used are of utmost importance. Furthermore, the best-performing models are far from achieving excellent levels of performance. Researchers should consider how to design better modeling techniques to effectively learn the knowledge embedded in textual data and contemplate how to prepare or annotate high-quality datasets.

## Acknowledgments

We would like to thank the following for their help: Oscar Jiang, Cathy Wang, David Lee, Mensur Wang and Steven Sun.

| | **bloomz_560m** | **bloomz_1b1** | **bloomz_3b** | **bloomz_7b1_mt** | **ChatGLM6B** | **MOSS 16B** | **GPT-3.5-turbo** |
|---|---|---|---|---|---|---|---|
| **Chinese** | 0.267 | 0.298 | 0.229 | 0.229 | 0.411 | 0.252 | 0.298 |
| **Math** | 0.251 | 0.233 | 0.251 | 0.281 | 0.343 | 0.278 | 0.427 |
| **Physics** | 0.167 | 0.185 | 0.179 | 0.220 | 0.208 | 0.220 | 0.333 |
| **Chemistry** | 0.300 | 0.300 | 0.380 | 0.220 | 0.360 | 0.280 | 0.360 |
| **Politics** | 0.256 | 0.244 | 0.332 | 0.381 | 0.228 | 0.258 | 0.469 |
| **History** | 0.264 | 0.256 | 0.399 | 0.397 | 0.413 | 0.289 | 0.532 |
| **Geography** | 0.255 | 0.248 | 0.316 | 0.349 | 0.328 | 0.064 | 0.477 |
| **Biology** | 0.274 | 0.278 | 0.354 | 0.350 | 0.338 | 0.173 | 0.620 |
| **Average** | 0.254 | 0.255 | 0.305 | 0.303 | 0.329 | 0.227 | 0.440 |

Table 4: Few-shot accuracy of all models on education subtasks